\documentclass[letterpaper, 10 pt, conference]{ieeeconf}  

\IEEEoverridecommandlockouts                              

\usepackage{graphicx}  
\usepackage{booktabs}
\usepackage{hyperref}

\title{\LARGE \bf
Longitudinal Digital Phenotyping for Early Cognitive-Motor Screening}

\author{Diego Jimenez-Oviedo$^{1}$, Ruben Vera-Rodriguez$^{1,*}$, Ruben Tolosana$^{1}$,\\ Juan Carlos Ruiz-Garcia$^{1}$, and Jaime Herreros-Rodriguez$^{2}$\\
{\footnotesize $^{1}$BiometricsAI, Universidad Autonoma de Madrid, 28049 Madrid, Spain}\\{\footnotesize $^{2}$Hospital Universitario Infanta Leonor, 28031 Madrid, Spain}\\{\footnotesize*Corresponding author: \href{mailto:ruben.vera@uam.es}{ruben.vera@uam.es}}}

\begin{document}

\maketitle
\thispagestyle{empty}
\pagestyle{empty}

\begin{abstract}

Early detection of atypical cognitive-motor development is critical for timely intervention, yet traditional assessments rely heavily on subjective, static evaluations. The integration of digital devices offers an opportunity for continuous, objective monitoring through digital biomarkers.  

In this work, we propose an AI-driven longitudinal framework to model developmental trajectories in children aged 18 months to 8 years. Using a dataset of tablet-based interactions collected over multiple academic years, we analyzed six cognitive-motor tasks (e.g., fine motor control, reaction time). We applied dimensionality reduction (t-SNE) and unsupervised clustering (K-Means++) to identify distinct developmental phenotypes and tracked individual transitions between these profiles over time.  

Our analysis reveals three distinct profiles: low, medium, and high performance. Crucially, longitudinal tracking highlights a high stability in the low-performance cluster ($>$90\% retention in early years), suggesting that early deficits tend to persist without intervention. Conversely, higher-performance clusters show greater variability, potentially reflecting engagement factors.

This study validates the use of unsupervised learning on touchscreen data to uncover heterogeneous developmental paths. The identified profiles serve as scalable, data-driven proxies for cognitive growth, offering a foundation for early screening tools and personalized pediatric interventions.

\end{abstract}
\vspace{3pt}

\section{INTRODUCTION}

Early detection of atypical cognitive development in children is crucial for implementing timely and effective educational or clinical interventions. However, current assessment approaches are often static, limited to isolated measurements, and heavily dependent on subjective evaluations. With the growing integration of digital devices in early childhood education, there is an opportunity to collect rich, objective, and continuous data about children's cognitive and motor behavior.

In this work, we present a longitudinal modeling framework that uses unsupervised learning to track cognitive development based on interaction data gathered from young children during structured tablet-based activities. Over several academic years, children completed a series of digital tasks designed to assess a range of cognitive-motor functions. The resulting performance data allow for quantitative profiling of developmental stages and trajectories.

We apply dimensionality reduction and clustering methods to segment children into developmental profiles and then analyze how these profiles evolve over time. Our goal is not to classify children into diagnostic categories, but rather to uncover naturally emerging patterns of cognitive growth and change. This approach enables a deeper understanding of developmental variability and provides a foundation for scalable, data-driven monitoring in early education and pediatric research.


\section{Related Work}

Longitudinal studies examining children's interaction with mobile devices have traditionally relied on indirect data sources such as parental questionnaires, structured interviews, and media diaries \cite{mcharg2020,vandewater2007}. While informative, these approaches often suffer from limited frequency of measurement and subjective bias.

Recent research has begun to incorporate automated data collection methods, offering more granular insights into cognitive and motor development. For instance, Radesky et al.~\cite{radesky2020young,radesky2023} and Pedersen et al.~\cite{pedersen2023} leveraged mobile interaction logs and embedded game-based assessments to evaluate executive functions and cognitive abilities in early childhood. These efforts demonstrate the potential of touch interaction data—such as gesture accuracy, timing, and error rates—as proxies for neuropsychological metrics.

Further advancements include the use of multimodal sensor data from tablets and wearable devices, enabling fine-grained analysis of pressure, movement, and physiological states during interaction \cite{baltru2019, suneesh2025}. Such approaches facilitate a more holistic understanding of child–device engagement and support the development of adaptive educational technologies.

However, most existing datasets cover limited age ranges or offer sparse temporal resolution. In contrast, this work builds upon the dense, automated, and sensor-based longitudinal ChildCIdb dataset \cite{tolosana2022tetc} covering children from 18 months to 8 years of age, with repeated measurements across multiple academic years.

\section{Methods}

\subsection{Dataset}

\begin{figure*}[t]
    \centering
    \includegraphics[width=0.75\textwidth]{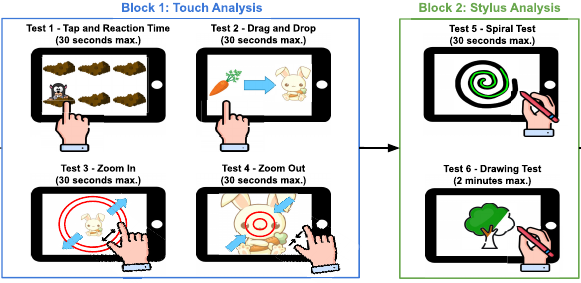}
    \caption{Overview of the longitudinal data collection and processing pipeline used in the study.}
    \label{fig:pipeline}
\end{figure*}

The dataset consists of longitudinal interaction data collected from children between the ages of 18 months and 8 years old, enrolled in an early education program at the school Las Suertes in Madrid, Spain. Each participant interacted with a tablet device during annual cognitive-motor assessment sessions spanning up to five academic years. The tests were conducted in a controlled setting using the same device model (Samsung Galaxy Tab A 10.1) and standardized instructions to ensure consistency across sessions. The cohort represents a general population of students from a regular education center, with no prior exclusion criteria based on neuro-developmental diagnoses, thus capturing the natural variability of cognitive-motor growth in a school setting.

The database comprises structured interaction data from over 940 children aged between 18 months and 8 years, grouped into seven educational levels (Groups 2 to 8) in alignment with the Spanish education system, as shown in table \ref{tab:educational_levels}.

\begin{table}[t]
\centering
\caption{Relationship between educational levels (courses) and age ranges.}
\label{tab:educational_levels}
\setlength{\tabcolsep}{8pt} 
\begin{tabular}{ll}
\toprule
\textbf{Educational Level} & \textbf{Age Range} \\
\midrule
Course 2 & 18 Months - 2 Years \\
Course 3 & 2-3 Years \\
Course 4 & 3-4 Years \\
Course 5 & 4-5 Years \\
Course 6 & 5-6 Years \\
Course 7 & 6-7 Years \\
Course 8 & 7-8 Years \\
\bottomrule
\end{tabular}
\end{table}

As shown in Figure~\ref{fig:pipeline}, each child completed six digital tasks from the \textit{ChildCIdb} database  \cite{tolosana2022tetc} designed to assess motor and cognitive abilities. Interaction data were automatically logged and converted into quantitative performance metrics, including time-to-completion, gesture accuracy, and error rates, which were used to compute normalized scores for each test (Q1–Q6).

Each academic year is treated as a distinct time point, and data are available for varying numbers of sessions per child.

\subsection{Data Preparation}

The performance was scored using the formulas established by Ruiz-Garcia et al. [1].~\cite{ruizgarcia2024access}. These formulas compute a Q score between 0 and 100 for each test, where 100 represents optimal performance. Each formula uses different task-specific features—such as reaction time, precision, or coverage—depending on the nature of the interaction. The resulting Q1–Q6 scores form the input features used for clustering and longitudinal analysis. 

\subsection{Dimensionality Reduction and Clustering}

To visualize and cluster the data, a two-step approach was applied:

\begin{enumerate}
    \item \textbf{t-SNE (t-distributed stochastic neighbor embedding)} was used to reduce the six-dimensional test score vector to two dimensions while preserving local structure. This facilitated interpretation and cluster separation.
    \item \textbf{K-Means++ clustering} was then applied on the t-SNE embeddings to partition participants into cognitive profiles for each academic year. The optimal number of clusters was determined using the elbow method.
\end{enumerate}

Clusters were interpreted as representing low, medium, and high levels of performance, based on their average test scores across dimensions.

Because t-SNE prioritizes local neighborhood preservation rather than global distance fidelity, the resulting clusters are interpreted as descriptive groupings of relative performance within each cohort, rather than strict metric partitions of the original feature space.

\subsection{Longitudinal Analysis}

Cognitive progression was studied by tracking how each participant transitioned between clusters across consecutive academic years. Transition matrices were computed to capture movement between performance levels (e.g., from low to medium cluster). These matrices were used to characterize developmental trajectories, including stable growth, improvement, stagnation, or decline.

This cluster-based tracking approach enables us to model heterogeneous cognitive development paths without requiring predefined outcome labels, making it suitable for early-stage screening and monitoring.
\begin{itemize}
    \item \textbf{Q1 – Tap and Reaction Time:} Tapping moving targets to measure reaction time, precision, and attention.
    \item \textbf{Q2 – Drag and Drop:} Dragging objects to targets, evaluating hand-eye coordination, movement control, and goal-directed planning.
    \item \textbf{Q3 – Zoom In:} Enlarging an image with a two-finger gesture to fit a boundary, assessing bimanual coordination and force modulation.
    \item \textbf{Q4 – Zoom Out:} Reducing an image using a pinch gesture, evaluating similar motor and perceptual skills.
    \item \textbf{Q5 – Spiral Test:} Tracing a spiral with a stylus within boundaries to measure precision, stability, and fine motor control.
    \item \textbf{Q6 – Drawing Test:} Coloring a figure within a time limit to reflect attention, spatial awareness, planning, and motor coordination.
\end{itemize}


\section{Cognitive Profiles per Cluster}

\begin{figure*}[t]
    \centering
    \includegraphics[width=0.65\textwidth]{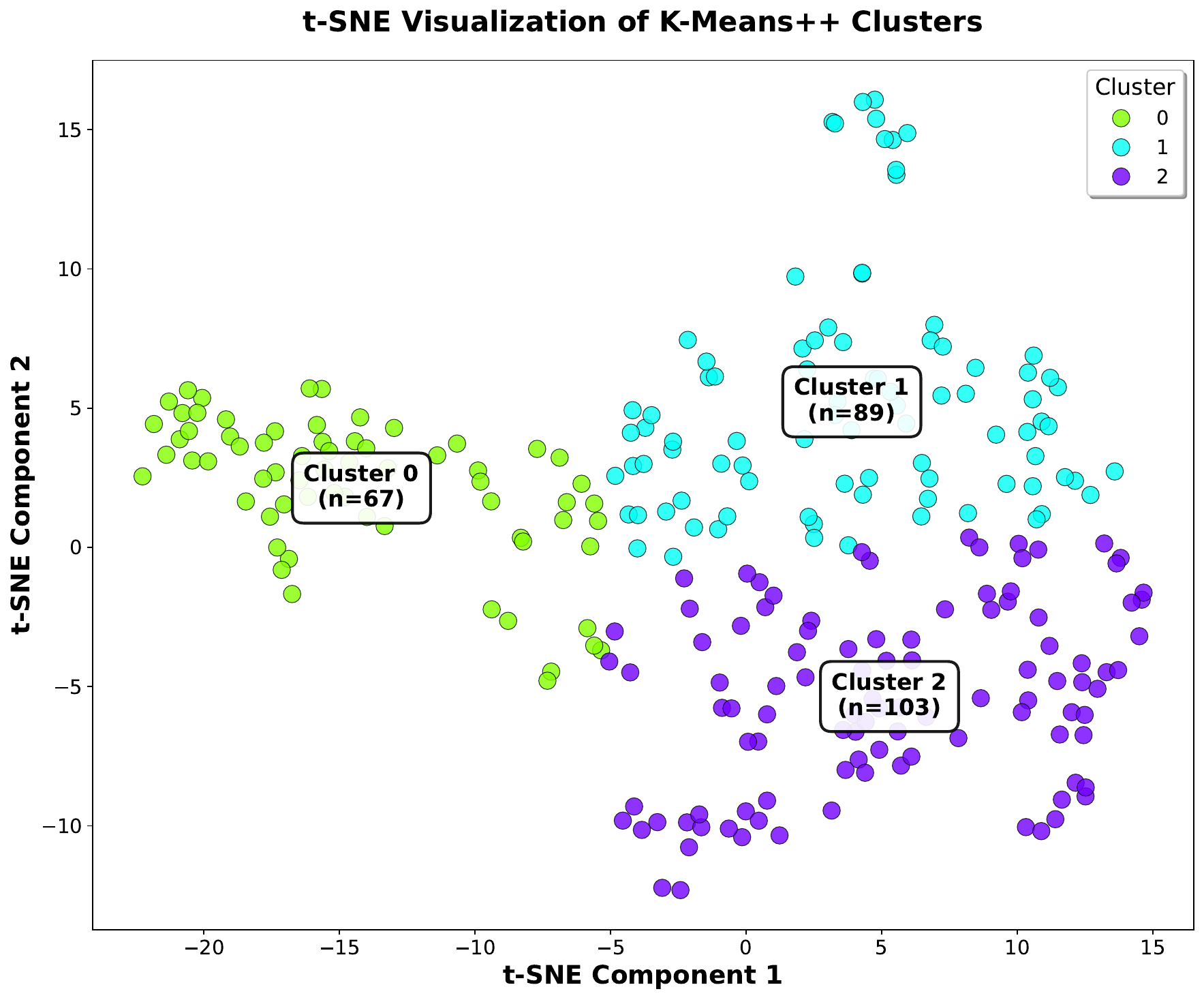}
    \caption{t-SNE projection of K-Means++ clustering results for Course~4 based on the six cognitive-motor task scores.}
    \label{fig:clustering_course4}
\end{figure*}

To better understand the underlying behavioral profiles of the participants, a clustering analysis was performed separately for each academic year. The clustering was based on the normalized scores obtained in the six tablet-based cognitive-motor tasks (Q1–Q6), allowing us to identify distinct patterns of performance within each cohort.

The number of clusters for each academic year was determined using the elbow method, which identifies the most appropriate number of clusters by analyzing the within-cluster variance. For students in courses 3 through 7, three clusters were found to best capture the variability in performance. In contrast, for courses 2 and 8, two clusters were sufficient. This pattern likely reflects the developmental and skill-related characteristics of the cohorts: course 2 students show more homogeneous performance due to limited experience and foundational skills, whereas course 8 students are generally more uniformly proficient.

As an illustrative example, Figure~\ref{fig:clustering_course4} shows a 2D projection of the clustering results for Course 4. This visualization was generated using dimensionality reduction (t-SNE) to plot the distribution of participants across clusters in a reduced feature space.

Following this example, Table~\ref{tab:clusters} shows the average scores obtained by children in each cluster across all six tasks (Q1–Q6), for every academic year from Course 2 to Course 8. The percentage column indicates the proportion of participants in each cluster within a given course.

\begin{table*}[t]
\centering
\caption{Average $Q$ scores per cluster and course. The percentage column indicates the distribution of samples in each cluster.}
\setlength{\tabcolsep}{4pt}
\begin{tabular}{ccccccccc}
\toprule
\textbf{Cluster} & \textbf{Q1} & \textbf{Q2} & \textbf{Q3} & \textbf{Q4} & \textbf{Q5} & \textbf{Q6} & \textbf{\%} & \textbf{Course} \\
\midrule
0 & 2.39 & 6.26 & 0.03 & 2.34 & 17.04 & 29.44 & 67.31 & 02 \\
1 & 52.99 & 5.07 & 0.00 & 3.81 & 16.51 & 20.42 & 32.69 & 02 \\
\midrule
0 & 2.82 & 5.86 & 0.50 & 2.00 & 16.34 & 28.04 & 40.80 & 03 \\
1 & 60.60 & 14.64 & 4.68 & 11.88 & 23.36 & 36.78 & 59.20 & 03 \\
\midrule
0 & 2.49 & 7.37 & 0.33 & 2.84 & 17.96 & 29.66 & 25.36 & 04 \\
1 & 63.20 & 15.77 & 1.21 & 8.86 & 26.43 & 42.90 & 40.71 & 04 \\
2 & 74.33 & 52.65 & 23.25 & 44.33 & 63.27 & 78.00 & 33.93 & 04 \\
\midrule
0 & 4.93 & 8.35 & 0.43 & 3.17 & 17.41 & 30.94 & 20.30 & 05 \\
1 & 69.92 & 34.46 & 6.20 & 20.97 & 31.24 & 58.25 & 38.01 & 05 \\
2 & 80.62 & 64.87 & 32.68 & 47.05 & 71.94 & 84.97 & 41.70 & 05 \\
\midrule
0 & 53.03 & 18.98 & 3.23 & 13.79 & 22.72 & 35.64 & 21.62 & 06 \\
1 & 74.20 & 74.86 & 17.29 & 28.41 & 58.64 & 89.13 & 38.61 & 06 \\
2 & 86.26 & 62.97 & 42.43 & 56.83 & 77.18 & 88.21 & 39.77 & 06 \\
\midrule
0 & 73.90 & 42.02 & 7.16 & 25.23 & 43.23 & 66.43 & 28.57 & 07 \\
1 & 73.15 & 83.89 & 30.77 & 37.76 & 43.77 & 89.15 & 27.62 & 07 \\
2 & 86.23 & 69.84 & 38.19 & 50.13 & 86.44 & 90.97 & 43.81 & 07 \\
\midrule
0 & 85.13 & 60.09 & 25.69 & 30.11 & 62.34 & 86.77 & 58.93 & 08 \\
1 & 78.41 & 88.39 & 31.53 & 52.10 & 77.35 & 91.77 & 41.07 & 08 \\
\bottomrule
\end{tabular}
\label{tab:clusters}
\end{table*}

\subsection{Cluster Interpretation and Cognitive Profiles}

Each cluster represents a distinct cognitive profile that reflects the variability in children's cognitive-motor abilities. These profiles provide valuable insights into their developmental stages and can help in tailoring educational interventions to better address their specific needs.

\begin{itemize}
    \item \textbf{Cluster 0 – Low performance:} Children in this cluster exhibit the lowest average scores across all six tasks. Their cognitive-motor profiles suggest difficulties in motor coordination, slower reaction times, and a lower ability to perform more complex gestures such as zooming, drawing, or multitouch tasks. These children tend to struggle with tasks that require fine motor control or rapid responses. This profile might indicate that the children are in the early stages of cognitive-motor development, where foundational skills are still being established. The overall low performance across all tasks suggests that these children may benefit from focused interventions that target basic motor and cognitive skills, helping them to improve coordination, reaction times, and task engagement.
    
    \item \textbf{Cluster 1 – Medium performance:} Children in this group show moderate scores, reflecting emerging cognitive-motor abilities. They perform better than those in Cluster 0 but still face challenges, particularly with tasks that require precise motor control, like multitouch gestures. These children demonstrate a developing ability to complete simpler tasks with greater ease, but they may struggle when tasks become more complex or require more advanced coordination. The medium performance suggests that these children are in a transitional phase, where their cognitive and motor skills are developing but may need additional support to handle increasingly difficult tasks. This group represents children who show promise but may require more focused attention to fully develop their cognitive-motor capabilities.
    
    \item \textbf{Cluster 2 – High performance:} This cluster consists of children with the highest scores across all tasks, indicating strong cognitive-motor abilities. These children are able to complete all six tasks with speed, accuracy, and confidence. Their high performance reflects advanced cognitive processing, rapid reaction times, and well-coordinated motor skills. These children excel at tasks requiring complex motor control, such as multitouch gestures, and demonstrate a strong understanding of task demands. This profile indicates that these children have reached an advanced stage in their cognitive-motor development, showing strong abilities in both simple and complex tasks. Their high scores suggest they are capable of handling more challenging cognitive and motor tasks, which makes them ideal candidates for enrichment programs aimed at further developing their skills.
\end{itemize}

These cognitive profiles are crucial for understanding the individual differences in children's cognitive-motor development. Identifying these profiles allows for a more targeted approach to education, where children can be provided with the appropriate level of challenge and support based on their specific needs. By recognizing the unique characteristics of each cluster, educators can adapt their teaching strategies to foster better outcomes for all children, regardless of their developmental stage.

\subsection{Condensed Analysis of Cluster Transitions}

To summarize the dynamics of developmental change over time, two tables are presented that capture both the overall transition patterns and the relative stability of each cognitive cluster.

\subsubsection*{Overall Cluster Transition Summary}

Table~\ref{tab:cluster_movement_summary} categorizes transitions between consecutive academic years into three types: \textbf{Stable}, \textbf{Improving}, and \textbf{Declining}. These percentages are based on the number of children who remained in, progressed to, or regressed from their original cluster.

\begin{table}[t]
\centering
\caption{Summary of cluster mobility across course transitions. Percentages indicate the proportion of children whose cluster status remained stable, improved, or declined.}
\label{tab:cluster_movement_summary}
\setlength{\tabcolsep}{6pt} 
\begin{tabular}{lccc}
\toprule
\textbf{Course Transition} & \textbf{Stable (\%)} & \textbf{Improving (\%)} & \textbf{Declining (\%)} \\
\midrule
Course 2 $\rightarrow$ 3 & 94.1 & 5.9 & 0.0 \\
Course 3 $\rightarrow$ 4 & 85.7 & 9.4 & 4.9 \\
Course 4 $\rightarrow$ 5 & 78.6 & 0.0 & 21.4 \\
Course 5 $\rightarrow$ 6 & 73.2 & 11.0 & 15.8 \\
Course 6 $\rightarrow$ 7 & 69.4 & 14.3 & 16.3 \\
Course 7 $\rightarrow$ 8 & 82.0 & 7.0 & 11.0 \\
\bottomrule
\end{tabular}
\end{table}

We observe a general trend of high stability, particularly in the early years (Courses 2–4). Declines are more prominent than improvements in most transitions, especially between Courses 4 and 5. This asymmetry highlights the inertia of cognitive profiles once established.

\subsubsection*{Cluster-Specific Stability Rates}

To further understand which cognitive profiles are most stable over time: Table~\ref{tab:cluster_stability} shows the proportion of children who remained in the same cluster from one year to the next, disaggregated by initial cluster.

\begin{table}[t]
\centering
\caption{Stability rates (\%) of each cluster across course transitions. High percentages reflect consistent developmental profiles year-over-year.}
\label{tab:cluster_stability}
\setlength{\tabcolsep}{6pt} 
\begin{tabular}{lccc}
\toprule
\textbf{Course Transition} & \textbf{Cluster 0} & \textbf{Cluster 1} & \textbf{Cluster 2} \\
\midrule
Course 2 $\rightarrow$ 3 & 100.0 & 88.0 & -- \\
Course 3 $\rightarrow$ 4 & 92.7 & 80.1 & 66.6 \\
Course 4 $\rightarrow$ 5 & 94.1 & 77.4 & 60.0 \\
Course 5 $\rightarrow$ 6 & 91.3 & 68.2 & 62.4 \\
Course 6 $\rightarrow$ 7 & 89.5 & 70.4 & 66.1 \\
Course 7 $\rightarrow$ 8 & 68.5 & 95.1 & -- \\
\bottomrule
\end{tabular}
\end{table}

\noindent
These results clearly show that:
\begin{itemize}
    \item \textbf{Cluster 0 (low performance)} is the most stable group across all years, often exceeding 90\% retention.
    \item \textbf{Cluster 1} exhibits moderate stability, but shows both upward and downward transitions.
    \item \textbf{Cluster 2 (high performance)} is the least stable group, with more children transitioning to lower clusters over time, particularly between Courses 4 and 6.
\end{itemize}

\section{Conclusion}

\subsection{Conclusion of Results}

Our analysis revealed important insights into the cognitive development of young children, particularly in terms of the stability and evolution of their developmental profiles. A key finding is the high stability observed in the low-performance cluster (Cluster 0), especially during the early academic years. The near 100\% stability from Course 2 to Course 3 and consistent high stability throughout subsequent transitions indicate that children in this group tend to maintain a relatively low level of performance over time. This finding has significant implications for early educational practices, as it suggests that children with low cognitive-motor performance at an early age may require more sustained and targeted interventions. The high stability within this cluster implies that, without early intervention, children may continue to struggle with foundational cognitive skills, which could impact their long-term academic success.

Conversely, the medium (Cluster 1) and high-performance (Cluster 2) clusters exhibit notable fluctuations, with a mix of stable, improving, and declining trajectories across the academic years. This suggests that while many children in these clusters demonstrate continuous growth, a subset may experience periods of stagnation or regression. These variations highlight the importance of personalized and adaptive educational strategies that cater to the diverse needs of children. By closely monitoring these transitions, educators and clinicians can identify children who may benefit from additional support or enrichment, ensuring that they remain on track for optimal cognitive development.

Additionally, the observed transition patterns between clusters underscore the heterogeneity of early cognitive development. While most children remain in their original cluster, some move between performance levels, indicating that cognitive growth is not a linear process. These findings further support the idea that personalized monitoring and interventions are crucial, as a one-size-fits-all approach may not be suitable for all children.

Regarding the transitions observed in high-performance clusters, the moderate decline in stability—particularly between Courses 4 and 6—does not necessarily imply a regression in neurodevelopmental capabilities. Instead, these fluctuations likely reflect the influence of non-cognitive factors, such as varying levels of task engagement or a ceiling effect in the metrics as children mature. This contrast further emphasizes the clinical relevance of Cluster 0, where the high persistence of low performance suggests a more stable developmental phenotype that may require prioritized screening and intervention

\subsection{Future Research}

Future research could expand upon our findings by incorporating other dimensions of child development, such as social and emotional growth, to better understand the factors influencing cognitive trajectories. Exploring how children's interactions with peers or caregivers affect their cognitive development could provide a more holistic view. Additionally, integrating more diverse data types, such as behavioral assessments or neuroimaging data, could offer deeper insights into the underlying mechanisms driving the observed developmental patterns.

Another interesting direction for future work is to examine the impact of specific interventions on the cognitive progression of children, particularly those in the low-performance clusters. Understanding which educational or clinical interventions are most effective at improving performance in these children could inform the development of targeted strategies for early intervention. It would also be valuable to investigate whether certain types of interventions lead to changes in the stability of cognitive profiles or facilitate transitions from low to medium or high-performance clusters.

Moreover, given the high stability observed in the low-performance cluster, future research could explore the effectiveness of adaptive and personalized educational tools. These tools could adjust the complexity of tasks or provide scaffolding based on the individual child's cognitive progress, potentially improving engagement and outcomes for children who might otherwise fall behind. 

Finally, extending this study to include a larger and more diverse sample of children across different cultural and socio-economic backgrounds would help determine whether the observed trends are generalizable. Investigating how context-specific factors influence cognitive development could enhance the applicability of this approach in real-world settings, such as schools or clinics, and ensure that early intervention strategies are tailored to the needs of various populations.

\addtolength{\textheight}{-12cm}   




\section*{ACKNOWLEDGMENT}



This project has been supported by PowerAI+ (SI4/PJI/2024- 00062 Comunidad de Madrid and UAM), Cátedra ENIA UAM-Veridas en IA Responsable (NextGenerationEU PRTR TSI-100927-2023-2), and TRUST-ID (PID2025-173396OB-I00 MICIU/AEI and the EU). Also, we acknowledge the computer resources provided by Centro de Computación Científica-Universidad Autónoma de Madrid (CCC-UAM).


\end{document}